RESEARCH ARTICLE

# Posttraumatic Stress Disorder Hyperarousal Event Detection Using Smartwatch Physiological and Activity Data

**Mahnoosh Sadeghi [1], Anthony D McDonald [1], and Farzan Sasangohar [1], \***

1. Department of Industrial and /systems Engineering, Texas A&M University;

m7979@tamu.edu; mcdonald@tamu.edu; sasangohar@tamu.edu

\* sasangohar@tamu.edu

## Abstract

Posttraumatic Stress Disorder (PTSD) is a psychiatric condition affecting nearly a quarter of the United States war veterans who return from war zones.  Treatment for PTSD typically consists of a combination of in-session therapy and medication. However; patients often experience their most severe PTSD symptoms outside of therapy sessions. Mobile health applications may address this gap, but their effectiveness is limited by the current gap in continuous monitoring and detection capabilities enabling timely intervention. The goal of this article is to develop a novel method to detect hyperarousal events using physiological and activity-based machine learning algorithms. Physiological data including heart rate and body acceleration as well as self-reported hyperarousal events were collected using a tool developed for commercial off-the-shelf wearable devices from 99 United States veterans diagnosed with PTSD over several days. The data were used to develop four machine learning algorithms: Random Forest, Support Vector Machine, Logistic Regression and XGBoost. The XGBoost model had the best performance in detecting onset of PTSD symptoms with over 83% accuracy and an AUC of 0.70. Post-hoc

SHapley Additive exPlanations (SHAP) additive explanation analysis showed that algorithm predictions were correlated with average heart rate, minimum heart rate and average body acceleration. Findings show promise in detecting onset of PTSD symptoms which could be the basis for developing remote and continuous monitoring systems for PTSD. Such systems may address a vital gap in just-in-time interventions for PTSD self-management outside of scheduled clinical appointments.

## Introduction

PTSD is a psychiatric condition experienced by individuals after exposure to life-threatening events, such as physical assault, sexual abuse, and combat exposure (1). PTSD symptomology includes avoidance, hyperarousal, and reexperiencing trauma through dreams and recollections (1). Avoidance symptoms include circumventing activities or thoughts associated with the traumatic event, decreased interest in daily life, and an overall feeling of detachment from one's surroundings. Hyperarousal symptoms include hypervigilance, feelings of irritability, and an exaggerated response following a startling event. Other symptoms of PTSD include anxiety, insomnia, fatigue, anger, and aggression (2,3).

Over 70% of the U.S. population will experience a traumatic event in their lifetime of whom 20% will subsequently develop Post-Traumatic Stress Disorder (PTSD) (4). Combat veterans are particularly prone to PTSD (5), with recent estimates of prevalence as high as 20% (4). Veterans with PTSD are also at a greater risk of suicide (6) , and suicidal thoughts (7)—according to a 2019 U.S. Department of Veteran Affairs annual report, an average of 20 veterans per day commit suicide, with a majority of cases linked to PTSD (8). Beyond personal costs, PTSD has an enormous societal cost associated with healthcare service utilization. The costs of caring for war veterans usually peaks 30-40 years following a major conflict. For example, for Iraq/Afghanistan conflicts, it is estimated that costs of veterans' care will peak around 2035 (9,10). This increased healthcare use is estimated to cost the United States over $60 billion each year and the costs are also expected to increase due to secondary and tertiary comorbidities including depression, substance abuse, smoking, heart disease, obesity, diabetes, chronic fatigue, and increased dementia (9,11).

PTSD is typically managed by a combination of therapeutic and pharmaceutical treatments, although many cases go undiagnosed or untreated potentially due to mental illness stigma and care shortages (12). Therapeutic methods include eye movement desensitization and reprocessing; exposure therapy; cognitive therapy; cognitive restructuring therapy; cognitive processing therapy; stress inoculation therapy as part of stress management therapy (13); and the trauma-focused cognitive behavioral therapy (14). While these methods are effective (15,16), there are several barriers to care access including geographical, financial, and cultural constraints, and limited care delivery resources (17,18). In addition, most intense symptoms of PTSD are often experienced outside clinical environments and in-between therapeutic sessions (19). Therefore, there is a critical need for tools and methods for real-time monitoring and detection of PTSD signs and symptoms, as well as mechanisms to support self-management of such symptoms. Recent advances in wearable physiological sensors and mobile health (mHealth) technologies may provide a viable alternative to address this need.

Prior work has shown that PTSD is correlated with several physiological measures including heart rate, heart rate variability, blood pressure, respiratory rate and skin conductance (20). Among these, heart rate has shown promise as a reliable correlate of PTSD (21). Sadeghi et al. (22) used the Autoregressive Integrated Moving Average (ARIMA) analysis to model veterans' heart rate patterns during PTSD hyperarousal events. Their results indicated strong correlation between heart rate characteristics (e.g., autocorrelation and fluctuation) and hyperarousal events. Recent efforts have utilized supervised machine learning tools to detect perceived stress using heart rate, other physiological metrics, and self-reported measures with reported accuracies ranging between 67%-92% (23–28), however, to our knowledge, only one study (23) used machine learning algorithms to predict the onset of PTSD symptoms among veterans based on heart rate data.

In their study, McDonald et al. (23) investigated self-reported periods of hyperarousal to extract heart rate time-dependent features and developed five machine learning algorithms: a Conventional Neural Network, Neural Network, Support Vector Machine (SVM), Random Forest, and Decision Tree. Among these methods, the SVM showed the highest accuracy (over 70%). While this study provided preliminary evidence supporting the efficacy of using heart rate to detect hyperarousal events, using heart data alone may be subject to significant noise associated with movement or physical activity (29,30). In line with previous research (e.g., 25,26,31), McDonald et al. (23) suggested that body acceleration data might improve the accuracy of machine learning algorithms and enable algorithms to distinguish between heart rate fluctuations due to physical activity and heart rate fluctuations due to mental stress. Therefore, the objective of this article is to expand McDonald et al.'s (23) study to machine learning algorithms that uses body acceleration and heart rate data to predict PTSD hyperarousal events in veterans. In addition, in an effort to improve the interpretation of the algorithm, we further analyze the developed model to investigate significant factors contributing to model's detection output.

## Materials and Methods

Four machine learning algorithms were trained using self-reported data collected naturalistically from veterans to predict PTSD hyperarousal events: Random Forest, XGBoost, Logistic Regression and non-linear SVM.

## Participants

Participants were recruited from seven Project Hero's United Healthcare Ride 2 Recovery (R2R) challenges. Project Hero is a non-profit organization dedicated to help veterans and first responders diagnosed with PTSD. In each challenge, veterans rode for an average of 7 days between key destinations in California, Washington DC, Minneapolis, Texas, and Nevada. Each day of the challenge involved approximately 8 hours of biking with the remaining time for resting and socializing. The research team joined a total of 5 rides in 2017, 2018 and 2019. Data from 99 veteran participants (82 male; 17 female) were used in this study. Participants' age ranged from 22 to 75 years old (M = 45.5, SD = 10). Majority of participants reported Veterans Affairs disability rating of over 90% related to PTSD. Table 1 summarizes other relevant demographics.

Table 1: participants' demographics; the numbers show the number of veterans per each group

| Ethnicity | | Branch | | VA Disability Ratings for PTSD | |
|---|---|---|---|---|---|
| American Indian or Alaska Native | 4 | Air Force | 5 | 40% | 2 |
| Asian | 1 | Army | 68 | 50% | 4 |
| Black/African American | 15 | 40% | 2 | 70% | 6 |
| Hispanic/Latino | 26 | 50% | 4 | 80% | 13 |
| Native Hawaiian | 1 | 70% | 6 | ≥ 90% | 74 |
| White | 44 | 80% | 13 | | |
| Other | 8 | ≥ 90% | 74 | | |

**Data Collection**

The data collection application (app) for smart wearable devices utilized in (23) was used. Participants were asked to wear smart watches (MOTO 360 Gen 1 or Gen 2, Apple Watch series 3 or 4) with the app installed on them for the duration of the study. The app ran continuously in the background and connected to participants' phones for the purpose of data transfer. The app had the ability to continuously and remotely collect physiological data including heart rate and acceleration from participants at the frequency of 1 Hz. The app included functionality which allowed the user to report a hyperarousal event (symptomatic of PTSD) through a simple 'double tap' anywhere on the watch face which created a time-stamped self-reported event. These events were used for training the machine learning algorithm.

**Data preprocessing**

All data analysis including data preprocessing and machine learning were conducted in Python 3.8.2 and R 3.6.2. The data preprocessing included four main steps: (1) imputation, (2) windowing and labeling, (3) dividing the data into training and testing, and (4) resampling the training dataset.

*Data Imputation*

Kalman filter imputation was used to impute missing acceleration and heart rate data. Kalman filter imputation is an established method for timeseries data imputation (32), especially for heart rate data (33). To determine the cut off range, we calculated the average Mean Square Error (MSE) of the imputed data and corresponding actual values. A cut off range of 15 MSE for estimating the randomly dropped values is suggested by Gui et al. (33). Based on Kalman filter imputation analysis, we chose 5 as the maximum imputation range because it was the greatest value among a set of successive values to have the highest MSE less than 15 (23).

*Windowing and labeling*

To investigate the patterns of hyperarousal events, the data was divided into 60-second sliding windows with 30 seconds overlap, chosen based on prior work (24) to predict stress severity based on physiological reactions. Each window was assigned a label based on the presence or absence of reported hyperarousal events. If a hyperarousal event occurred anywhere in the window, it was labeled as hyperarousal event; otherwise, it was labeled as non-hyperarousal event. All windows with over 80% missing values were dropped from the dataset. The final dataset included 530 and 13,554 instances of hyperarousal and non-hyperarousal events, respectively.

*Training, testing, and upsampling*

To validate the algorithm, the data was separated into training (70%) and testing (30%) sets by participant. Participants were assigned to only the testing or training set to ensure generalizability of the results. Table 2 shows the initial dataset classifications. One of the challenges of training the algorithm to detect PTSD hyperarousal events was the imbalanced dataset—96.2% of the windows were labeled non-hyperarousal events. To address this issue, we upsampled the training data. Upsampling was used because it decreases the information lost in the quantification process, thereby reducing the noise and increasing the resolution of the results (34,35). Upsampling has several advantages over downsampling. The main advantage of upsampling is its ability to use the entire information in the data rather than omitting part of the data (36). In addition, for noisy datasets, which is often the case when collecting data in naturalistic settings, oversampling is more robust to the noise in the data than undersampling and performs better for prediction (37). Based on a sensitivity analysis comparing different resampling ratio including 1-1, 2-1, 3-1, 3-2, and 4-3, a ratio of 4 (non-hyperarousal events) to 3 (hyperarousal events) windows was used for upsampling (Table 2).

Table 2: Training and testing datasets after and before resampling

|  | Label | Training set | Test set |
|---|---|---|---|
| Training and testing datasets before resampling | Non-hyperarousal events | 9486 | 4068 |
|  | Hyperarousal events | 372 | 158 |
| Training and testing datasets after resampling | Non-hyperarousal events | 9486 | 4068 |
|  | Hyperarousal events | 7114 | 158 |

## Feature Generation and Selection

*Heart rate data*

Previous research has shown that that time domain features of heart rate are strongly correlated with PTSD (38). We extracted time domain features of heart rate including maximum heart rate (bpm), minimum heart rate (bpm), heart rate standard deviation (bpm), heart rate range (max-min) (bpm), and average heart rate (bpm) from each window of time to use for PTSD hyperarousal prediction. Key features were extracted based on recommendations from a review article on detecting psychological stress using bio signals (39).

*Acceleration Data*

Research on stress prediction have used scalars of body acceleration to estimate body activity and to remove noise from the data (25,39–41). Garcia-Ceja et al. (40) used time domain and frequency domain features of acceleration to predict stress in participants in real work environments. In line with these approaches, we calculated the vector of body acceleration for each moment using the following widely used formula: body acceleration = $\sqrt{a_x^2 + a_y^2 + a_z^2}$ Where $a_x$ is the body acceleration in X direction, $a_y$ is body acceleration in Y direction, and $a_z$ is body acceleration in Z direction.

Further, based on previous research (e.g., 25,40), time domain features of body acceleration including average body acceleration (m/s$^2$), maximum body acceleration (m/s$^2$), minimum body acceleration (m/s$^2$), and range of body acceleration (m/s$^2$) were extracted to feed machine learning algorithms.

*Model Assessment*

Four models including Random Forest (maximum depth = 28, size of feature set = 10, number of trees = 50), XGBoost (Maximum depth = 37, number of trees = 50, learning rate = 0.3), Logistic

Regression (Lambda =0 .03240) and non-linear SVM (C=5, Sigma =12, Radial Basis Function) were trained to predict hyperarousal events. We then generated a confusion matrix to assess the performance of each model. Model comparisons were conducted with a 5x2 cross validation test following the recommendations in Dietterich (42) to minimize the type 1 error. This method uses p values and t statistics to compare the algorithms. The null hypothesis indicates that there is no significant difference between the algorithms in terms of performance (average accuracy) where the alternative hypothesis shows that one algorithm is more accurate than the others. Algorithms were further assessed with the Area Under the receiver operating characteristic (ROC) Curve (AUC).

*Feature Importance and Model Interpretation*

The complexity of black box machine learning models and the need to make these models explainable necessitate an evaluation of the influence of algorithm's features on algorithm predictions. In this study, we used Shapley Additive exPlanations (SHAP) to address this. SHAP uses game theoretic concepts to allocate values to features in a model based on their importance in prediction (43). SHAP values indicate how much each feature contributes to the prediction of the machine learning algorithm. SHAP value summary plots generate a feature importance list along with the distribution of each feature and shows how each value affects the output of the model. This method is computationally efficient, consistence with human intuition, and interpretable for explaining class differences (44). Using SHAP values to interpret machine learning algorithms has several advantages over using more traditional methods such as dependency plots (44). For example, dependence plots do not usually show features' distributions, which may lead to misinterpreting regions with significant missing data. Conversely, SHAP plots show feature distributions. SHAP values also indicate how much a feature affects the output of the prediction by considering interaction effects, whereas partial dependence plots do not account for interactions between features.

## Results

### Model Performance and Comparison

Figure 1 shows the ROC curves for each algorithm along with the AUC values. As shown in the plots, XGBoost had the highest AUC (0.70). Random Forest, Logistic Regression, and non-linear Kernel SVM had AUC of 0.63, 0.62, and 0.61, respectively. Table 3 shows confusion matrices for developed algorithms at three probability cutoffs. The first confusion matrix prioritizes hyperarousal detection, the second confusion matrix balances the true positives and false positives rate, and the third matrix prioritizes minimizing false positive rates. The pairwise 5*2 cross validation test results showed that XGBoost significantly outperformed the Random Forest ($t = -13.25$, $p < 0.001$), SVM ($t = -13.02$, $p < 0.001$), and Logistic Regression ($t = -11.97$, $p < 0.001$). Based on the results from this table, XGBoost showed the best performance in detecting PTSD hyperarousal events.

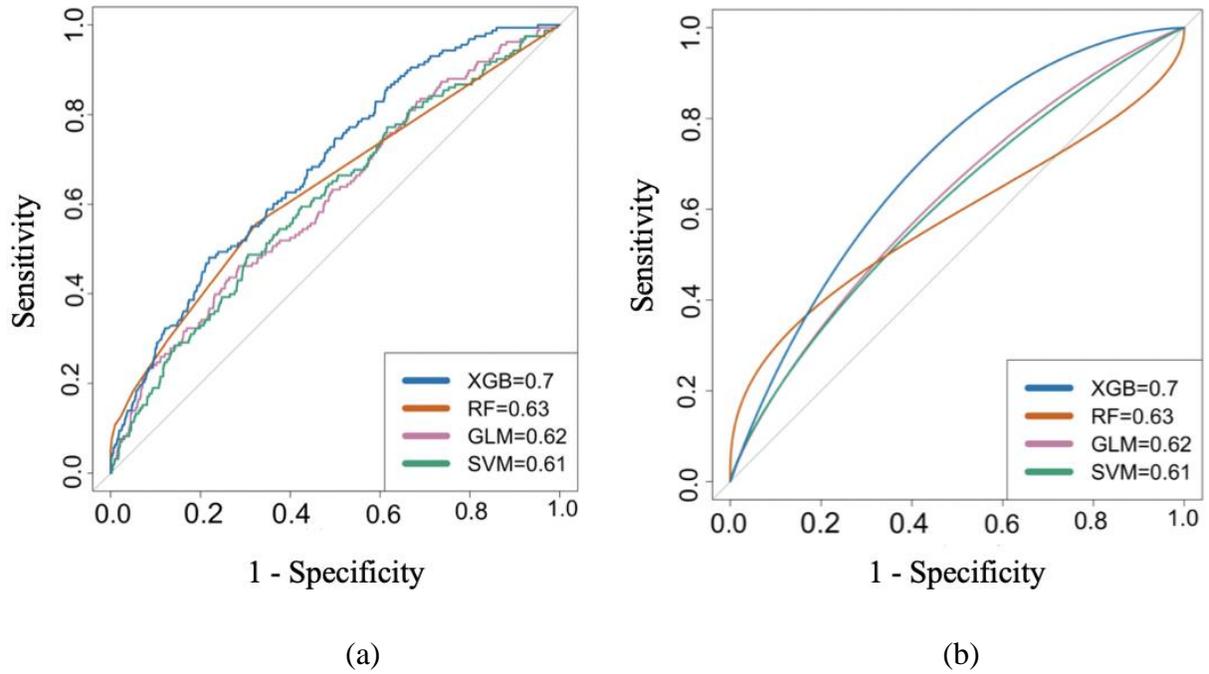

(a) (b)

Figure 1. AUC-ROC (a) empirical and (b) smooth (down) curves for algorithm

Table 3. Confusion matrices for all models at different probability cut offs.

| Design Performance | Algorithm | TP | FN | FP | TN | TPR | FPR |
|---|---|---|---|---|---|---|---|
| Prioritize hyperarousal detection (TPR=1) | XGB | 158 | 0 | 3821 | 247 | 1 | 0.94 |
|  | RF | 158 | 0 | 3853 | 205 | 1 | 0.95 |
|  | GLM | 158 | 0 | 4066 | 2 | 1 | 0.99 |
|  | SVM | 158 | 0 | 4065 | 3 | 1 | 0.99 |
| Balanced priorities (TPR=0.5) | XGB | 79 | 79 | 1064 | 3004 | 0.5 | 0.26 |
|  | RF | 88 | 70 | 1322 | 2746 | 0.55 | 0.33 |
|  | GLM | 79 | 79 | 1467 | 2601 | 0.5 | 0.36 |
|  | SVM | 79 | 79 | 1401 | 2667 | 0.5 | 0.34 |
|  | XGB | 46 | 112 | 420 | 3648 | 0.29 | 0.1 |
| Prioritize false positive minimization (FPR=0.1) | RF | 29 | 129 | 205 | 3863 | 0.18 | 0.1 |
| | GLM | 38 | 120 | 410 | 3658 | 0.24 | 0.1 |
| | SVM | 30 | 128 | 409 | 3659 | 0.19 | 0.1 |

**Model Interpretation**

Figure 2 shows the SHAP summary plot for the XGBoost model. In this figure, the Y-axis shows the feature as well as the mean SHAP values ordered from top to bottom, the color shows the significance of the feature's value in predicting the output, and the X-axis indicates how the feature affects the output of the model (whether that feature with that specific value is contributing to experiencing a hyperarousal event or not). The X axis further indicates log-odds of perceiving a PTSD hyperarousal event. According to the SHAP analysis, the most important body acceleration features are average body acceleration (linaccmean) and minimum body acceleration (linaccmin). The most important heart rate time-domain features for predicting PTSD hyperarousal events are minimum heart rate (hrmin) and heart rate standard deviation (hrsd).

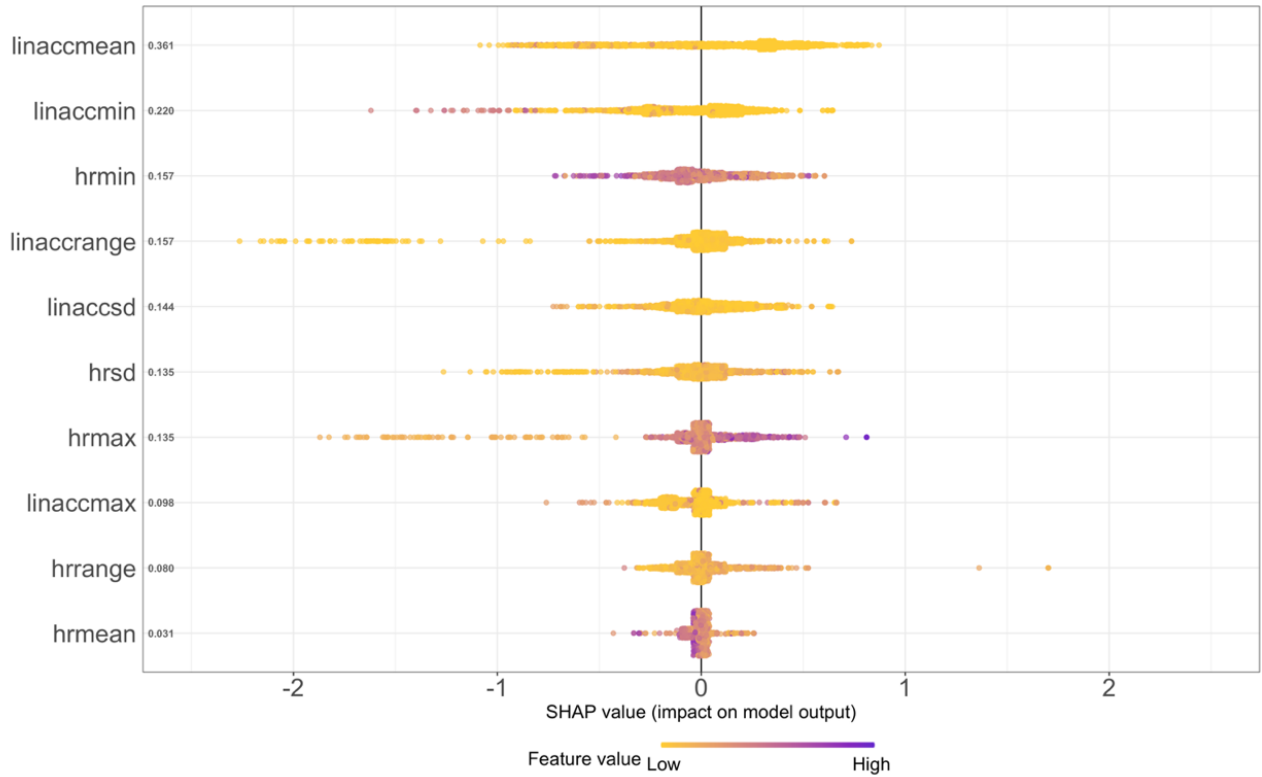

Figure 2. SHAP summary plot, Y axis shows each of the variables, and X axis shows log odds of perceiving a hyperarousal event.

Figure 3 shows the SHAP dependence plots for the two most important acceleration and heart rate features. SHAP dependence plots show contribution of a specific feature to a model based on the feature's distribution. In this plot each point shows an observation from the dataset, the X-axis line shows the value of the feature in that row, and the Y-axis shows the SHAP value for that feature that indicates the effect of that feature with that specific value on the prediction. The unit of X-axis is the same as the unit of the feature (for instance for heart rate measures it is beats per minute), and the unit of the Y-axis is log-odds of perceiving a PTSD hyperarousal event.

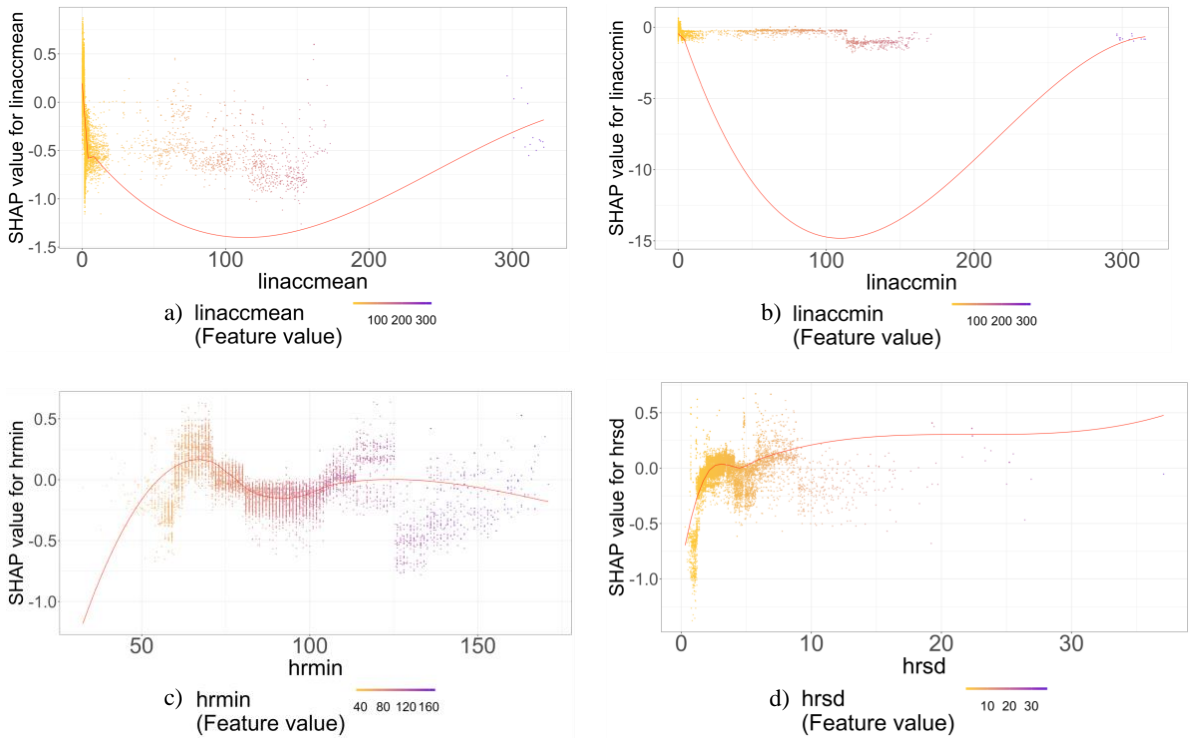

Figure 3. SHAP dependence plots, a) SHAP plot for average body acceleration (m/s$^2$), b) SHAP plot for minimum body acceleration (m/s$^2$), c) SHAP plot for minimum heart rate (bpm), and d) SHAP plot for heart rate standard deviation (bpm)

As shown in Figure 4, hyperarousal events are more likely to be observed with higher minimum heart rate values over the window. When the minimum heart rate is over 140 the risk of perceiving hyperarousal events increases. Also, as the average body acceleration and minimum body acceleration increase, the odds of the detecting PTSD hyperarousal events decrease. Finally, higher heart rate standard deviation, i.e. higher heart rate fluctuation, increases the risk of hyperarousal events.

## Discussion

This study developed, evaluated, and explicated machine learning algorithms to predict PTSD hyperarousal events among veterans using smartwatch based naturalistic heart rate and accelerometer data. The ground truth was subjectively-reported PTSD hyperarousal events. After preprocessing the data, we trained four different algorithms including Random Forest, SVM, Logistic Regression and XGBoost. Among the developed algorithms, the XGBoost was the most robust algorithm which yielded an AUC of 0.70 and over 81% accuracy. We sorted the most important features in the prediction process. The top three body acceleration features included average body acceleration, minimum body acceleration and range of body acceleration. The top three heart rate time domain features were minimum heart rate, standard deviation of heart rate and maximum heart rate. The initial analysis from the SHAP summary plot and SHAP dependence plots show that heart rate and body acceleration features have nonlinear relationships with PTSD episodes. A deeper look into SHAP plots indicate that as the body acceleration increases, indicating more activity from the participant, the algorithm is less likely to predict a PTSD hyperarousal events. This result is consistent with prior studies demonstrating a significant relationship between increased physical activity and a reduction in PTSD hyperarousal events (45–47).

The SHAP dependence plot for the average heart rate data corroborates that when the heart rate is between 60-70 bpm, PTSD hyperarousal events are more likely to happen (cf. 48). The SHAP summary plot indicates that heart rate standard deviation was one of the most important features contributing to the odds that the algorithm will predict PTSD hyperarousal event manifestation. In particular, our findings suggest that as the heart rate standard deviation increases, i.e., as heart rate fluctuates more and in higher ranges, the odds of detecting a PTSD hyperarousal event increases. This result supports previous findings (22,49,50) which showed that during PTSD hyperarousal events, participants experience increased heart rate acceleration and fluctuation.

This study builds on a prior study conducted by McDonald et al. (23) but there are several key differences that both enhance the algorithm and improve our understanding of algorithm performance. First, we expanded McDonald et al.'s dataset by conducting two additional field studies. Second, McDonald et al. indicated that one of the main limitations of their work is utilizing downsampling for data preprocessing. We addressed this limitation by using an upsampling method for preprocessing the data. Third, McDonald et al. used frequency domain features of heart rate such as coefficients of Fourier decomposition of heart rate. In our analysis we used time domain features of heart rate which may provide more tangible and interpretable results for this specific context (38,51). This change may be beneficial for integrating the detection algorithm into a treatment device because explaining predictions to veterans based on time domain features (e.g., standard deviation of heart rate) will be more understandable than frequency domain features (e.g., the phase of the 5th Fourier component). Fourth, in our analysis we added body acceleration features to decrease the noise in data by differentiating heart rate changes due to physical activity versus fluctuations due to stress (similar to (27)) and used a new XGBoost algorithms to train the machine learning model. We believe these changes have contributed to the significant improvement in the machine learning algorithm performance compared to the findings from McDonald et al. (81% in this study compared to 70 % in (23)). Finally, McDonald et al explain that one of their study limitations is the limited insight into the performance-shaping factors for the algorithm predictions. We addressed this issue by adding feature importance and SHAP analysis to improve the interpretability of the results.

Several limitations of this study should be addressed in future work. First, stress and PTSD hyperarousal events are highly idiosyncratic. A stimulus that triggers one individual may or may not trigger someone else. Because of the subjective and sustained characteristics of stress, defining the start, end, duration, and intensity of a hyperarousal event is an uncertain task (52). As a result, it is significantly complex and difficult to define and measure a ground truth for stress. Hyperarousal events might have been over or under reported due to the subjectivity of the perceived events. Individual differences such as gender, age, lifestyle and other factors can affect PTSD hyperarousal events; therefore, personalizing machine learning algorithms might boost their performance. Another issue in this study was the high number of heart rate missing values due to the naturalistic nature of the study. While non-intrusiveness of smart watches makes them suitable for naturalistic data collection, like most wrist-based sensors, smart watches use optical technology the accuracy of which is affected by skin tone and proximity to skin (53). Future work may validate the findings presented here using more accurate sensors such as chest straps that use electrical pulse. Lastly, although machine learning algorithms work in theory, external validation of these algorithms in naturalistic settings are necessary to evaluate the accuracy and applicability of these algorithms in the real world settings.

This article provides preliminary evidence of efficacy for data-driven real-time PTSD hyperarousal detection tools that can be used beyond clinic walls to remotely and continuously monitor veterans suffering from PTSD. In addition to the promise shown by the machine learning algorithm, in this article we utilized analytical techniques to which identifies most important features contributing to such detection, hence, improving the interpretation of the outcomes and moving towards explainable ML tools for PTSD monitoring. Although other machine learning algorithms exist to detect stress, to the best of our knowledge, the algorithm documented in this paper is one of very few algorithms that is specific to PTSD. The work is in progress to validate this algorithm in longitudinal home studies using smart watches and smart phones.